\documentclass[runningheads]{llncs}
\usepackage{graphicx}

\usepackage{comment}
\usepackage{amsmath,amssymb} 
\usepackage{color}
\usepackage{booktabs}
\usepackage{cite}
\usepackage{multirow}
\usepackage{enumitem}
\usepackage{floatrow}
\usepackage{wrapfig}
\usepackage[pagebackref,breaklinks,colorlinks, citecolor=black]{hyperref}
\usepackage[accsupp]{axessibility}  

\newcommand{\etal}{\textit{et al}.}
\newcommand{\ie}{\textit{i}.\textit{e}.}
\newcommand{\eg}{\textit{e}.\textit{g}.}

\begin{document}
\pagestyle{headings}
\mainmatter
\def\ECCVSubNumber{****}  

\title{Domain Knowledge-Informed Self-Supervised Representations for Workout Form Assessment} 

\titlerunning{Domain Knowledge-Informed Self-Supervised Learning}
%
\author{Paritosh Parmar\inst{1,2} \and Amol Gharat\inst{2} \and
Helge Rhodin\inst{1}}
\authorrunning{P. Parmar et al.}
%
\institute{University of British Columbia \and
FlexAI Inc.}
\maketitle
\begin{abstract}
Maintaining proper form while exercising is important for preventing injuries and maximizing muscle mass gains. Detecting errors in workout form naturally requires estimating human’s body pose. However, off-the-shelf pose estimators struggle to perform well on the videos recorded in gym scenarios due to factors such as camera angles, occlusion from gym equipment, illumination, and clothing. To aggravate the problem, the errors to be detected in the workouts are very subtle. To that end, we propose to learn exercise-oriented image and video representations from unlabeled samples such that a small dataset annotated by experts suffices for supervised error detection. In particular, our domain knowledge-informed self-supervised approaches (pose contrastive learning and motion disentangling) exploit the harmonic motion of the exercise actions, and capitalize on the large variances in camera angles, clothes, and illumination to learn powerful representations. To facilitate our self-supervised pretraining, and supervised finetuning, we curated a new exercise dataset, \emph{Fitness-AQA} (\url{https://github.com/ParitoshParmar/Fitness-AQA}), comprising of three exercises: BackSquat, BarbellRow, and OverheadPress. It has been  annotated by expert trainers for multiple crucial and typically occurring exercise errors. Experimental results show that our self-supervised representations outperform off-the-shelf 2D- and 3D-pose estimators and several other baselines. We also show that our approaches can be applied to other domains/tasks such as pose estimation and dive quality assessment.
\end{abstract}
\section{Introduction}
\label{sec:introduction}
Detecting errors in gym exercise execution and providing feedback on it is crucial for preventing injuries and maximizing muscle gain. However, feedback from personal trainers is a costly option and hence used only sparingly\textemdash typically only a few days a month, just enough to learn the basic form. We believe that an automated computer vision-based workout form assessment (\eg, in the form of an app) would provide a cheap and viable substitute for personal trainers to continuously monitor users’ workout form when their trainers are not around. Such an option would also be helpful to the socio-economically disadvantaged demographic who cannot afford or have access to personal trainers.

While fitness apps have recently become popular, the existing apps only allow the users to make workout plans\textemdash they do not provide a functionality to assess the workout form of the users. To detect errors in the workout videos, it is important to analyze the posture of the person. Academic research in workout form assessment so far has been limited to simple, controlled conditions \cite{ogata, chen2020pose}, where posture can be reliably estimated using off-the-shelf (OTS) pose estimators \cite{openpose,hmr,spin}. Ours, on the other hand, is the first work to tackle the problem of workout form assessment distinctly in complex, real-world gym scenarios, where, people generally record themselves using ubiquitous cellphone cameras that they place somewhere in the vicinity; which results in large variances in terms of camera angles, alongside clothing styles, lighting, and occlusions due to gym equipment (barbells, dumbbells, racks). These environmental factors combined with the subtle nature of workout errors (refer to Fig.~\ref{fig:ots_fails}) and the convoluted, uncommon poses that people go through while exercising, cause major challenges for OTS pose estimators (refer to Fig.~\ref{fig:ots_fails}), and consequently, workout form errors cannot be reliably detected from pose. To mitigate this in the absence of workout datasets labeled for human body pose, we propose to replace the error-prone pose estimators with our more robust domain knowledge-informed self-supervised representations that are sensitive to pose and motion, learned from unlabeled videos --- helps in avoiding annotation efforts. Towards those ends, our contributions are as follows:
\begin{figure}
\floatbox[{\capbeside\thisfloatsetup{capbesideposition={right,center},capbesidewidth=0.55\textwidth}}]{figure}[\FBwidth]
{\caption{\textbf{Concept.} (a) Errors of small magnitude generally occurring in workout form: Good column shows correct posture/execution (knees should be outwards), while the Bad column shows erroneous form during exercising. (b) Examples of failures of off-the-shelf 2D- and 3D-pose estimators in real-world gym scenarios (compare the discrepancies in pose estimation with the magnitude of the errors to be detected). We tackle the problem of detecting errors in workout form. To do so more accurately, we replace the error-prone pose estimators with our more robust fitness domain-oriented representations learnt using self-supervision.}}
{\includegraphics[width=0.4\textwidth]{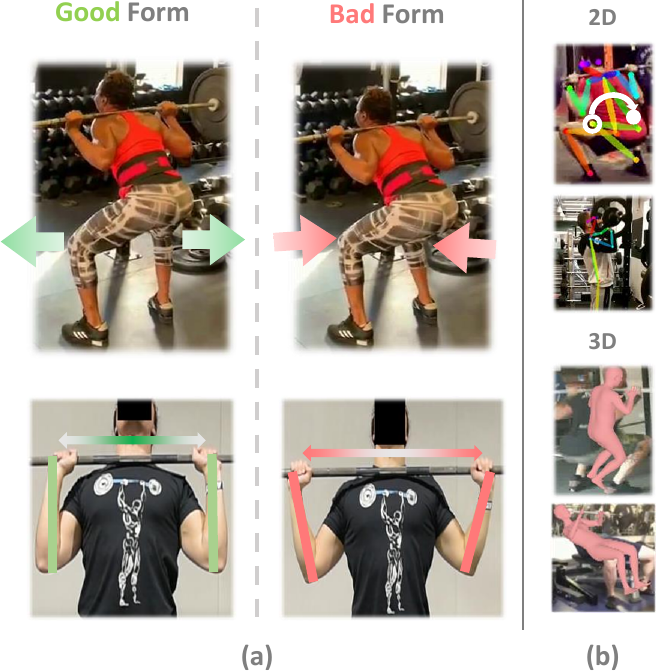}}
\label{fig:ots_fails}
\end{figure}

\begin{enumerate}[noitemsep, leftmargin=*]
\item \textbf{Novel self-supervised approaches that leverage domain knowledge}. We initiate the work in the direction of domain knowledge-informed self-supervised representation learning by developing two contrastive learning-based approaches that capitalize on the harmonic motion of workout actions and the large variance in unlabeled gym videos to learn robust fitness domain-oriented representations (Sec.~\ref{sec:approach}). Our domain knowledge-informed self-supervised representations outperform 2D- and 3D-pose estimators \cite{openpose, spin, ogata}, and various general self-supervised approaches \cite{simsiam, benaim2020speednet, video_rotnet, jenni2020video} on the task of workout form assessment on existing and our newly introduced datasets. This indicates that future work on representation learning would benefit from using domain knowledge in designing self-supervised methods, especially when tackling problems involving real-world data.
\item \textbf{Workout form assessment dataset}. To facilitate our self-supervised approaches, as well as the subsequent supervised workout form error detection, we collected the largest, first-of-its-kind, in-the-wild, fine-grained fitness assessment dataset, covering three different exercises (Sec.~\ref{sec:dataset}) and a small labeled subset for evaluation. 
We show that this in-the-wild dataset provides a significantly more challenging benchmark than the existing ones recorded in controlled conditions.
\end{enumerate}
\begin{figure}
\floatbox[{\capbeside\thisfloatsetup{capbesideposition={right,center},capbesidewidth=0.45\textwidth}}]{figure}[\FBwidth]
{\caption{\textbf{Fitness-AQA dataset hierarchy.} Numbers below the dataset type indicate dataset size; and those under the errors indicate the ratio of non-erroneous:erroneous samples. I, V indicate if the error detection is static image- or multiframe (video)-based.}}
{\includegraphics[width=0.5\textwidth]{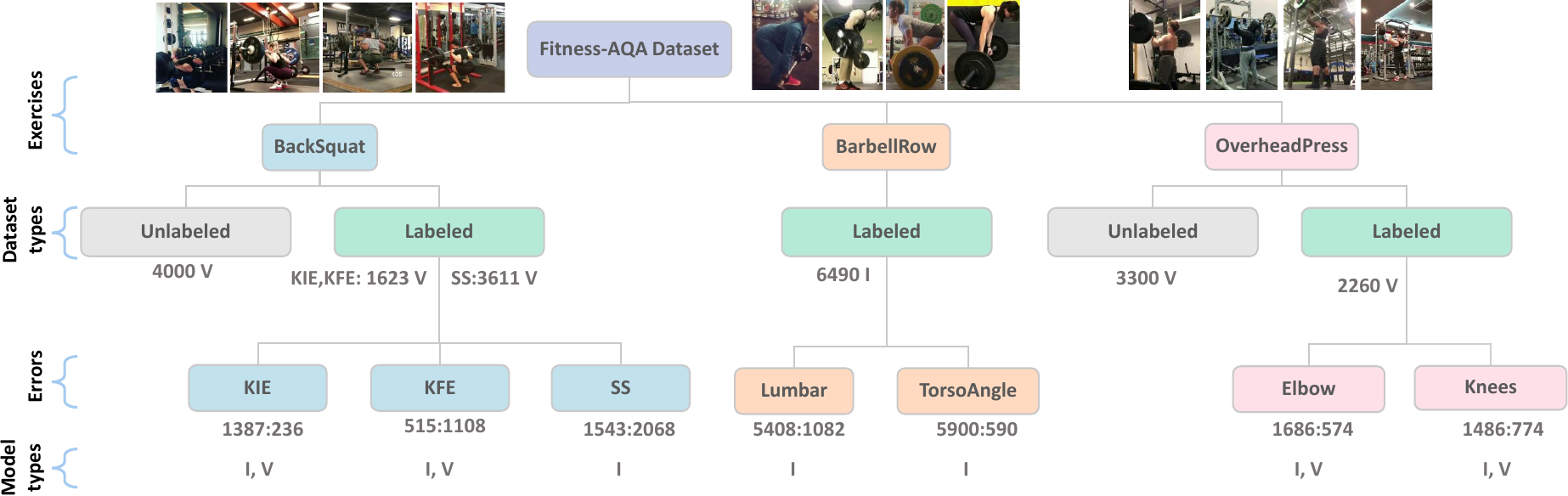}}
\label{fig:dataset_fitness-aqa}
\end{figure}
\section{Related Work}
\paragraph{Action Quality Assessment (AQA)/Skills Assessment (SA).} 
Our work can be classified under AQA/SA, which involve the computer vision-based quantification of the quality of movements and actions. Works in AQA/SA have mainly been focused on domains like physiotherapy \cite{tao2016comparative, parmar2016measuring, li2021improving, sardari2020vi, du2021assessing}, Olympic sports \cite{parmar2019action, mtlaqa, xu2019learning, tang2020uncertainty, yu2021group, chen2021sportscap}, various types of skills \cite{doughty2019pros, wang2020towards, liu2021towards, parmar2021piano}. However, workout form assessment, especially, in real-world conditions, has not received much attention.

Approaches in AQA can be organized into 1) human pose features-based \cite{pirsiavash, Pan_2019_ICCV}; 2) image and video features-based \cite{parmar2017learning, mtlaqa}. Pose-based approaches use OTS pose estimators to extract 2D or 3D coordinate positions of various human body joints. These approaches have the disadvantage that poor estimation of the pose can adversely affect the final output. This is especially prevalent in non-daily action classes like fitness and sports domains. This can be mitigated, for example, by annotating domain-specific datasets \cite{chen2021sportscap}, but that requires a considerable amount of manual annotation efforts, financial resources, and 3D annotations can only be obtained in controlled conditions. Therefore, we propose to learn domain-oriented pose-sensitive representations from unlabeled videos, which can be finetuned using only a small labeled dataset.

Closest to ours is the work on backsquat assessment by Ogata \etal \cite{ogata}. However, a) they used OTS pose estimators, whereas we develop self-supervised approaches to learn more powerful representations; b) being dependent on OTS pose estimators, their approach is limited only to simple, controlled environments, whereas our approach is applicable to complex, real-world scenarios (Sec.~\ref{sec:experiments}); and c) their dataset contains only single exercise and was collected in simpler conditions and a single human, whereas our dataset contains three exercises and was collected in real-world gym scenarios and numerous humans (further differences discussed in Sec.~\ref{sec:dataset}).
\paragraph{SSL.} 
Earlier work in this area include those of autoencoders \cite{hinton2006reducing}, which learn low-dimensional representations by reconstructing the input. Le \etal \cite{convisa} propose a way to learn hierarchical representations from unlabeled videos using unsupervised learning, which was also considered as a feature extractor in an earlier AQA work \cite{pirsiavash}, but was found to perform worse than an OTS pose estimator. Recently, Chen \etal \cite{simsiam} proposed a simple siamese approach to learn representations that obtain competitive results on various benchmarks. Various general SSL works also propose to leverage properties of videos. Misra \etal \cite{misra2016shuffle} and Xu \etal \cite{xu2019self} propose to exploit temporal order of frames and clips. Predicting the amount of rotation  in images and videos was used as a pretext task by Gidaris \etal \cite{rotnet} and Jing \etal \cite{video_rotnet}. Wang \etal \cite{wang2019self} leveraged motion and appearance statistics to learn self-supervised video representations. Benaim \etal \cite{benaim2020speednet} and Wang \etal \cite{wang2020self} used video speed prediction as the pretext task. In addition to video speed prediction, Jenni \etal \cite{jenni2020video} proposed to use wider range of temporal transformations for pretext task. In contrast, we developed domain knowledge-informed SSL approaches that we show outperform general SSL approaches. A few works propose to leverage time-contrast to learn representations using self-supervision \cite{hyvarinen2016unsupervised, tcn, honari2020unsupervised}. However, these temporal models either consider a single-view or a single subject. Our pose contrastive approach, on the other hand, simultaneously exploits cross-view and cross-subject information to learn more meaningful representations. 

Another work proposes to disentangle pose and appearance from multiple views with a geometry-aware representation \cite{rhodin2018unsupervised}. However, this approach is not tailored for exercise analysis, and requires calibrated multi-view datasets. Inspired by this method, we develop a variant\textemdash our pose and appearance disentangling baseline\textemdash applicable to our dataset.
\section{Method}
\label{sec:approach}
\begin{figure}
\floatbox[{\capbeside\thisfloatsetup{capbesideposition={right,center},capbesidewidth=0.4\textwidth}}]{figure}[\FBwidth]
{\caption{\textbf{Barbell trajectory.} \textcolor{red}{Red bounding boxes} (bboxes) - barbell object detected; \textcolor{red}{Red dots}: the center of bboxes; \textcolor{blue}{Blue} curve: the parabolic trajectory of the barbell traced out.}}
{\includegraphics[width=0.5\textwidth]{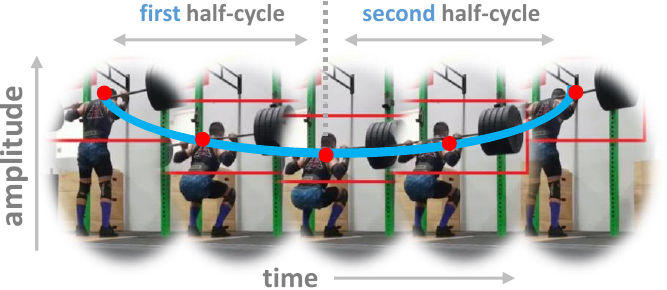}}
\label{fig:barbell_trajectory}
\end{figure}
 In this section, we present our self-supervised approaches for learning image and video representations. Subsequently, an error detection network is trained to map these self-supervised representations to workout form error probabilities. Note that in the following, we have presented our approaches using BackSquat as an exemplary exercise, but our methods are applicable to other exercises.\\
 
\noindent\textbf{Preliminary: \textit{Synchronizing videos}.}
 Our methods build upon quasi-synchro- nized videos. In some datasets, such as Human3.6M \cite{h36m_pami}, synchronized videos recorded from multiple angles is already available using special setups, which allows unsupervised learning, \eg, as done in \cite{rhodin2018unsupervised}. However, we are not using any kind of such special setups. Thus, we quasi-synchronize the videos using the following method. Given a collection of videos of (different) people performing the same exercise, we detect the barbell/weight over time to get a motion trajectory, which when plotted against time traces an approximately parabolic curve as shown in Fig.~\ref{fig:barbell_trajectory}. These trajectories are then amplitude-normalized. Object size, resolutions do not affect the normalization, as we are using the center of the bounding box; and the vertical movement of the barbell can be reliably recorded from various viewpoints (unless extreme, like top-view of the scene\textemdash unrealistic, anyway). Now, we leverage the following property to synchronize the videos: for a given elevation of the object (or equivalently, the amplitude of the trajectory), the people doing the same exercise would be in approximately the same pose. This holds across different subjects, different video instances, and across different views/camera angles, which allows us to synchronize videos of different subjects in different environments/scenes.
\subsection{Self-Supervised Pose Contrastive Learning}
\label{sec:approach_cvcspc}
\noindent\textbf{Objective.}
Given the synchronized video samples of the same exercise (\eg, BackSquat), in this approach, we aim to learn richer human pose information using self-supervised contrastive learning. In contrastive learning, same or similar samples are pulled together, while dissimilar samples are pushed apart \cite{chopra2005_contrastive}. In our case, we hypothesize that we can extend contrastive learning to learn human pose-sensitive representations. Particularly, we propose a self-supervised pretext task, which aims to pull together images (frames of videos) containing humans in similar poses, while pushing apart images with humans in dissimilar poses as shown in Fig.~\textcolor{red}{4}
. Note that, this approach operates on single frame-triplets (not videos or clips) at a time.
\begin{figure}
\floatbox[{\capbeside\thisfloatsetup{capbesideposition={right,center},capbesidewidth=0.6\textwidth}}]{figure}[\FBwidth]
{\caption{\textbf{Cross-View Cross-Subject Pose Contrastive learning (CVCSPC)}. \textcolor{red}{Red} lines indicate repulsion, while the \textcolor{green}{green} line indicates attraction in the representation space.}}
{\includegraphics[width=0.3\textwidth]{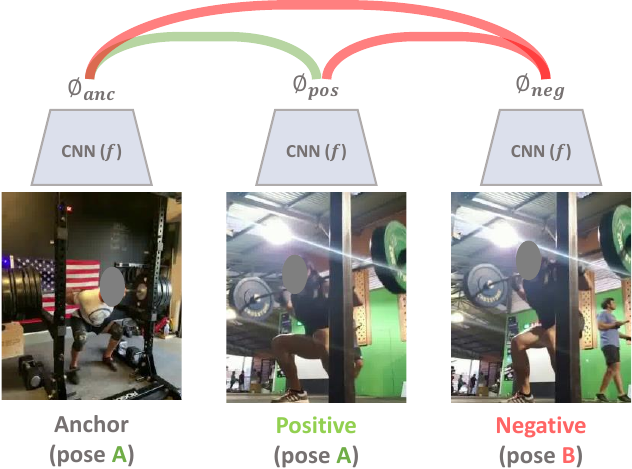}}
\label{fig:approach_pose_contrastive}
\end{figure}

\noindent\textbf{Constructing triplets for contrastive learning.}
 Once we have the normalized barbell trajectories, for any given anchor input, $I_{\text{anc}}$,  we retrieve the corresponding positive input frames with similar object elevation, $I_{\text{pos}}$, and the negative input frames with a difference in object elevation of more than a threshold value ($\delta$), $I_{\text{neg}}$, from across video instances; and subsequently build triplets of $\{I_{\text{anc}}, I_{\text{pos}}, I_{\text{neg}}\}$. Such triplets provide a cross-view, cross-subject, cross-video-instance self-supervisory signal that has not yet been leveraged by the existing computer vision approaches to learn pose sensitive representations. These triplets also offer strong, in-built data augmentations. A recent work \cite{ryali2021} observed that background augmentation can help increase the robustness of self-supervised learning. Our method not only provides such background augmentation, but also provides foreground augmentation in terms of appearance (clothing, body type, gender, etc.). We term our approach Cross-View Cross-Subject Pose Contrastive learning (CVCSPC). 

\noindent\textbf{Contrastive learning.} 
We use the constructed triplet, $\{I_{\text{anc}}, I_{\text{pos}}, I_{\text{neg}}\}$, to learn good representations through self-supervised contrastive learning. Let $f$ represent a 2D-convolutional neural network (CNN) backbone, which when applied to $I_{\text{anc}}, I_{\text{pos}}, I_{\text{neg}}$, yields $\phi_{\text{anc}}, \phi_{\text{pos}}, \phi_{\text{neg}}$, respectively. In contrastive learning, $\phi_{\text{anc}}$ and $\phi_{\text{pos}}$ are forced to be similar, \ie, $\phi_{\text{anc}} \approx \phi_{\text{pos}}$, while $\phi_{\text{anc}}$ and $\phi_{\text{neg}}$ are forced to be dissimilar, \ie, $\phi_{\text{anc}} \neq \phi_{\text{neg}}$, as illustrated in Fig.~\textcolor{red}{4}
. Following \cite{sigurdsson2018actor}, we optimize the parameters of $f$ during the self-supervised training, by minimizing the distance ratio loss \cite{hoffer2015deep},
\begin{equation}
\label{eq:distance_ratio_loss}
    \mathcal{L} = -\log \frac{\mathrm{e}^{-||\phi_{\text{anc}} - \phi_{\text{pos}}||_{2}}}{\mathrm{e}^{-||\phi_{\text{anc}} - \phi_{\text{pos}}||_{2}} + \mathrm{e}^{-||\phi_{\text{anc}} - \phi_{\text{neg}}||_{2}}}.
\end{equation}
\subsection{Self-Supervised Motion Disentangling}
\label{sec:approach_md}
Motion cues can be useful in detecting many workout form errors. Different from our pose-contrastive approach, this approach uses motion information to detect anomalies in workout form. In the following, we first present the preliminary information, before describing our method.

\begin{figure}
    \centering
    \includegraphics[width=0.5\textwidth]{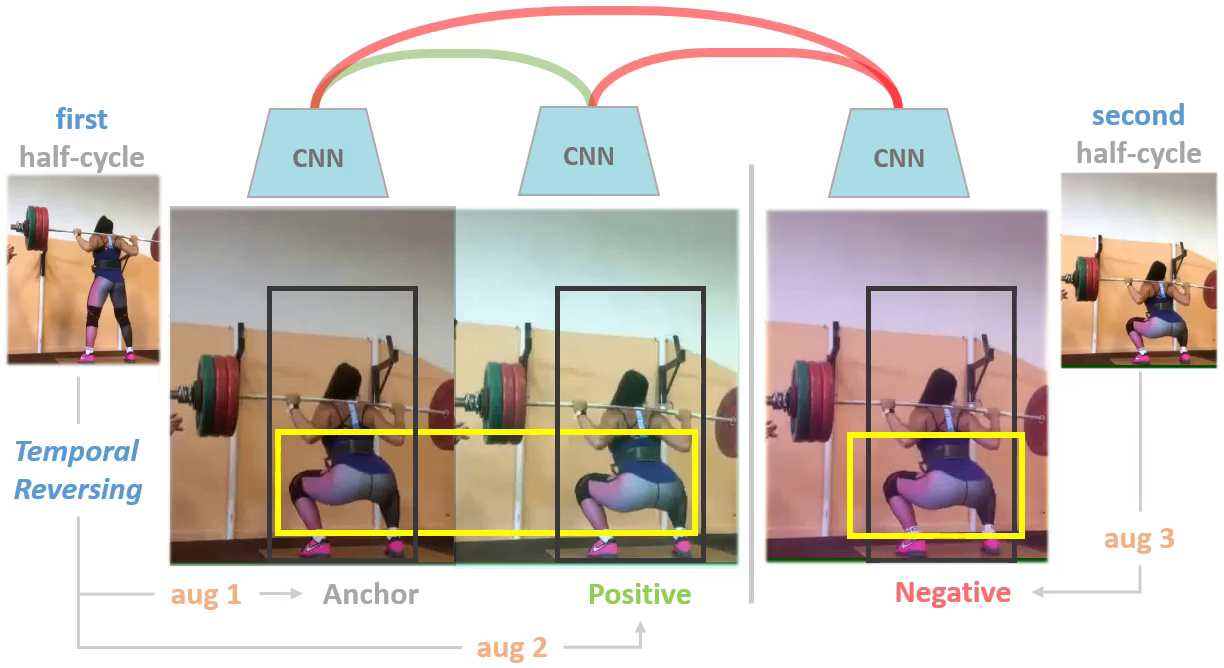}
    \caption{\textbf{Motion Disentangling (MD) approach.} \textit{Please view in AdobeReader to play the embedded animation for better explanation.} \colorbox{black}{\textcolor{white}{\textbf{Black}}} boxes: global motion (getting up, here); \colorbox{yellow}{\textbf{Yellow}} boxes: local motions (the knees rotating inwards under the influence of heavy training weight); aug: augmentations. Here we have applied very weak augmentation (only color augmentation) for representative purpose\textemdash to better illustrate the concept. However, in practice, we apply much stronger augmentations. \textcolor{red}{Red} lines indicate repulsion, while \textcolor{green}{green} indicates attraction in the representation space.}
    \label{fig:approach_md}
\end{figure}

\noindent\textbf{Preliminaries}
\begin{itemize}[noitemsep, topsep=0pt, leftmargin=*]
    \item \textbf{Useful property 1: Harmonic motion}. Workout actions have a desirable property of exhibiting harmonic motion. For example, during benchpress (an exercise targeting the chest muscles), the person would be lifting the barbell above their chest and then bringing it down to the starting point; or during squats, the person would be squatting down (first half-cycle in Fig.~\ref{fig:barbell_trajectory}) and then getting up (second half-cycle in Fig.~\ref{fig:barbell_trajectory}). 

    \item \textbf{Useful property 2: Bias in temporal location of form-errors}. People are more likely to make errors (anomalous motions) when lifting up the weights (one half-cycle of the harmonic motion, as in Fig.~\ref{fig:barbell_trajectory}), rather than lowering the weights (another half-cycle of the harmonic motion). 
    
    \item \textbf{Global motion.} The actual, regular motion of the workout action. For example, in Backsquat, the person squatting down and getting up.

    \item \textbf{Local motion.} The small-scale, fine-grained, irregular motion of the body parts (ref. Fig.~\ref{fig:approach_md}). For example, in Backsquat, the knees abnormally going inward/outward or forward. So, while the global motion refers to regularities in motion patterns, local motion would cover anomalies in motion patterns.
\end{itemize}
\noindent\textbf{Objective.}
Our goal is to learn self-supervised representations that are sensitive to local (anomalous) motions. The above discussed properties can provide a very useful, freely available signal that has not yet been exploited for this task by the existing computer vision approaches. We design a contrastive learning-based self-supervised approach to disentangle the local motion from the global motion.

\noindent\textbf{Accentuating the local motion.}
Temporally reversing any one of the half-cycles would, in general, make both half-cycles identical in terms of the global motion, while they would still differ in terms of the local motion. In other words, contrasting the two half-cycles after temporally reversing any one of them, helps accentuate the anomalous local motion, as shown in Fig.~\ref{fig:approach_md}.

\noindent\textbf{Constructing triplets for contrastive learning.} 
The first half-cycle serves as the anchor; an augmented copy of the anchor serves as the positive input. The second half-cycle serves as the negative input. As discussed previously, we randomly temporally-reverse either the \{anchor, positive\} pair or the \{negative\} input to make the global motion of all three identical. In practice, we randomly and independently applied the following augmentations on the triplets: image horizontal flipping, partial image masking, image translation, image rotation, image blurring, image zooming, color channel swapping, temporal shifting.

\noindent\textbf{Contrastive learning.}
We use a 3DCNN as the backbone for this model, and Eq.~\ref{eq:distance_ratio_loss} as the loss function for this self-supervision task. Through contrastive learning, the 3DCNN learns to capture the previously discussed local, anomalous motions that are accentuated in our specially created triplets.

Anomalous motions maybe harmful or they can be beneficial. For example, knees buckling inwards during squatting is harmful, while knees going outwards is not. Therefore, during the finetuning phase, we aim to calibrate representations learnt using self-supervision to distinguish between harmful irregularities and harmless variations.
\section{Fitness-AQA Dataset}
\label{sec:dataset}
Since exercise or workout assessment is an emerging field, there is a shortage of dedicated video datasets. To the best of our knowledge, the Waseda backsquat dataset by Ogata \etal  \cite{ogata} is the only publicly available such dataset. However, this dataset has shortcomings such as: it contains samples from a single human subject; the human subject is deliberately faking exercise errors; no kind of exercising weights, such as barbells and dumbbells, are used; the videos do not include realistic occlusions. 

\noindent \textit{Dataset Collection.} To fill the void of real-world datasets, we collected the largest exercise assessment dataset from video sharing sites such as Instagram and YouTube. We considered the following three exercises: 1) BackSquat; 2) BarbellRow; and 3) Overhead (shoulder) Press. In addition to the labeled data, we also collected an unlabeled dataset to learn human pose focused representations in self-supervised ways (discussed in Sec.~\ref{sec:approach}). The purpose of the labeled dataset is to finetune our models to do actual error detection and quantify the performance of our models. We have provided statistics and illustrated the full hierarchy of our Fitness-AQA dataset in Fig.~\textcolor{red}{2}
. Illustrations of exercise errors are provided in the supplementary material.

\noindent \textit{Annotations by Expert Trainers}. We employed two professional gym trainers to annotate our dataset for error labels. Due to this, even very subtle errors are caught and annotated accordingly. Errors range from very subtle to very severe.\\
Unique properties of our dataset:
\begin{itemize}[noitemsep, topsep=0pt, leftmargin=*]
    \item \textbf{Real-world videos}. Unlike the existing dataset \cite{ogata}, we collected our dataset from actual real-world videos in actual gyms recorded by the people without any scripts. Due to this, the videos are naturally recorded from a wide range of azimuthal angles, inclination angles, and distances. Our samples were automatically processed to contain a single repetition.
    
    \item \textbf{People making errors under the impact of actual weights}. In the existing dataset \cite{ogata}, people are instructed to make deliberate exercise mistakes without being under the influence of actual weights. Our dataset, on the other hand, captures cases where people are naturally making mistakes (without any instructions), under the influence of actually heavy weights. Due to this, we believe that there is no bias towards exaggerated errors, and contains natural, subtler error cases.
    
    \item \textbf{Occlusions}. Having captured in actual gyms, human subjects are partially occluded by barbell weights, weight racks or other equipment like benches. 
    
    \item \textbf{Various types of clothing, background, illumination}. Since we did not hire any specific group of people to collect the dataset, the samples in our dataset are likely to come from numerous unique individuals, which results in a large number of clothing styles, and colors; different gyms (in terms of the room arrangement, and the background); other people in the background; and lighting conditions.
    
    \item \textbf{Unusual poses}. Exercise actions result in much more convoluted human body positions than those covered in the existing pose estimation datasets.
\end{itemize}
\section{Experiments}
\label{sec:experiments}
To validate our contributions, we compared our features against various baselines and off-the-shelf pose estimators in simple (Case Study 1) and complex conditions (Case Study 2), showing significant improvements in the latter case. 

We took a two-step approach towards detecting errors in exercising videos. Our models were first trained on the unlabeled datasets using self-supervision, and then used as feature extractors on the supervised datasets. For imbalanced datasets, we used class weights (in cross-entropy loss) inversely proportional to the class size. Note that the labeled dataset contains only the exercise error as ground-truth annotation and no information related to human pose. As such, our models did not use any pose-related ground-truth.

For the motion disentangling model, since the temporal model is already baked in it, we simply finetuned the model end-to-end on the labeled dataset for error detection. We used 32 frames for all types of errors.

For all 2DCNN-based approaches, we a learnt ResNet1D temporal model \cite{ogata} that aggregates frame-level features for supervised error prediction on our labeled dataset. We used about 200 frames during error detection. Finetuning end-to-end on such a long sequence is not recommended \cite{parmar2017learning, xu2019learning, zeng2020hybrid}. Therefore, in this case, the 2DCNN backbone is not finetuned unless specified otherwise.\\
\noindent\textit{Implementation details.} 
We used ResNet-18 \cite{resnet} as the backbone CNN unless specified otherwise. We used custom YOLOv3 \cite{yolov3} to detect barbells/weights; and normalized the amplitudes of the trajectories to -180 to 180 (simply for a resemblance to a circle). Specifications regarding each approach are as follows:
\begin{itemize}[noitemsep, topsep=0pt, leftmargin=*]
    \item \textbf{Pose Contrastive approach (CVCSPC)}. We used a threshold gap of 30 between anchor/positive and negative inputs. We initialized our backbone CNN with ImageNet weights. We used ADAM optimizer \cite{adam} with an initial learning rate of 1e-4 and optimized for 100 epochs with a batch size of 25. 
    
    \item \textbf{Motion Disentanglement approach (MD)}. We used R(2+1)D-18 \cite{kensho} as our backbone CNN. We sampled 16 frames from each half-cycle. We randomly applied strong augmentations. We initialized our backbone CNN with Kinetics \cite{kinetics} pretrained weights. We optimized our models using ADAM optimizer with an initial learning rate of 1e-4 for 20 epochs with a batch size of 5.  
\end{itemize}
Further details provided in the supplementary material.
\subsection{Case Study 1: Simple Conditions}
\begin{table}
\setlength{\tabcolsep}{4pt}
\centering
\resizebox{0.4\textwidth}{!}{
\begin{tabular}{@{}lccccccc@{}}
\toprule
\multirow{2}{*}{\textbf{\begin{tabular}[c]{@{}l@{}}Feature\\ extraction\end{tabular}}} & \multirow{2}{*}{\textbf{Modality}} & \multicolumn{6}{c}{\textbf{Accuracies $\uparrow$}}                                                  \\ \cmidrule(l){3-8} 
                                                                                       &                                    & \textbf{KIE} & \textbf{CVRB} & \textbf{CCRB} & \textbf{SS} & \textbf{KFE} & \textbf{Avg} \\ \midrule
HMR-TDM \cite{ogata} & 3D Pose    & 89.80          & \textbf{98.65} & 93.05          & \textbf{87.30} & 83.58          & 89.08          \\
Ours CVCSPC & Image & \textbf{95.92} & 91.89          & \textbf{94.44} & 77.77          & \textbf{89.55} & \textbf{89.92} \\ \bottomrule
\end{tabular}
\caption{\textbf{Performance comparison on Waseda Squat dataset}.}
}
\label{tab:res_waseda}
\end{table}
The Waseda Squat dataset \cite{ogata} provides an excellent labeled dataset for evaluating exercise errors in controlled conditions. The publicly available portion of this dataset contains samples from a single human subject. This dataset was not captured in a gym-like setting, but rather in home, and office-like settings. Each sample contains multiple squat repetitions. Note that the publicly available train/val/test split is different from that used in the original paper. Using this dataset, we experimented detecting the following errors: knees inward error (KIE); convex rounded back (spine) (CVRB); concave rounded back (spine) (CCRB); shallow squat (SS); knees forward error (KFE). To do so, we trained classifiers to distinguish between each of these error classes and good squat class (samples belonging to this class did not contain any errors). In this experiment, we compared features from our CVCSPC method (self-supervisedly trained on our unlabeled BackSquat dataset) against the Temporal Distances Matrices (TDM) derived from HMR pose estimator \cite{hmr}. HMR-TDM features were made available by Ogata \etal \cite{ogata}. During feature extraction, we resized the input images to 320$\times$320 pixels, and considered the center 224$\times$224 pixel crop. We did not consider our MD model because this dataset has multiple repetitions in each sample, and the sequence length is 300 frames, which is about 9 times longer than our MD model sequence length (32 frames). And, consequently, if we temporally downsample the sequence, it would lose a lot of information.

The results are summarized in Table~\textcolor{red}{1}, where we report accuracies. We found that our model outperformed existing methods \cite{ogata} on three types of errors: KIE, CCRB, and KFE; with the performances being notably better on KIE and KFE errors. Even though not consistently across all the errors, our self-supervisedly learnt features outperformed HMR-TDM features on overall average performance. Note that large performance gap is not expected on this dataset, as OTS pose estimators work quite well in these simpler conditions.

\subsection{Case Study 2: In-The-Wild Conditions}
Next, we considered evaluating our approach on more complex datasets. For that, we considered our labeled datasets, which we introduced in Sec. \ref{sec:dataset}, where we also discussed the reasons that make our new in-the-wild dataset more challenging. Unless mentioned otherwise, we divided the datasets into train-, validation-, and test-splits of 70\%, 15\%, and 15\%, respectively.\\
\noindent\textit{Baselines.} 
We compared our self-supervised feature extractors with the following models and features:
\begin{itemize}[noitemsep, topsep=0pt, leftmargin=*]
    \item ImageNet: ImageNet \cite{imagenet} pretrained ResNet-18 \cite{resnet}
    \item Kinetics: Kinetics \cite{kinetics} pretrained R(2+1)D-18 \cite{kensho}
    \item SPIN-TDM: Temporal Distance Matrices (TDM) \cite{ogata} constructed from the output of SPIN \cite{spin} (3D joint positions)
    \item OpenPose-TDM: Temporal Distance Matrices \cite{ogata} constructed from the output of OpenPose \cite{openpose} (2D joint positions). Originally, TDM was proposed for 3D joint positions, but we also experiment with constructing TDMs from 2D joint positions. 
    \item SimSiam: ImageNet pretrained model adapted/trained to our dataset using a general self-supervised image representation learning approach: SimSiam \cite{simsiam}.
    \item Ours PAD: Inspired from \cite{rhodin2018unsupervised, park2020swapping}, we developed an autoencoder-based approach that learns to disentangle pose and appearance of the human. Pose vector is then used for error-detection. We term this pose and appearance disentangling approach Ours PAD. We initialized the encoder with ImageNet weights. We have elaborated on this baseline in the supplementary material.
    \item VideoSpeed-1: Kinetics pretrained model adapted/trained to our dataset using the pretext task of predicting speed of videos \cite{benaim2020speednet}. We considered the following speeds: 1x (normal), 2x (faster), 3x, 4x (fastest).
    \item VideoSpeed-2: same as VideoSpeed-1, but for 1x speed, we sampled frames uniformly from entire sequence. For higher speeds, it would create the effect of repeating the sequences. So, it can equivalently be considered as counting the exercise repetitions.
    \item VideoRot: Kinetics pretrained model adapted/trained to our dataset using the pretext task of predicting rotation amount of videos \cite{video_rotnet}. Rotation amount is selected randomly from \{0$^{\circ}$, 90$^{\circ}$, 180$^{\circ}$, 270$^{\circ}$\}.
    \item TemporalXform: Kinetics pretrained model adapted to our dataset using the pretext task of predicting various temporal transforms \cite{jenni2020video}.
    \item Ours TemporalXform-1: We developed a contrastive learning-based approach in which the negative input is more temporally shifted than the positive input relative to the anchor. We initialized with Kinetics pretrained model.
    \item Ours TemporalXform-2: We developed another contrastive learning-based approach in which the negative is more temporally distorted than the positive input. We initialized with Kinetics pretrained model.
\end{itemize}
\noindent\textit{Performance metric.} Since this dataset is imbalanced, we report the F1-score, instead of the accuracy.

\noindent\textbf{Dataset: Fitness-AQA BackSquat}

\noindent\textit{Knees Inward and Knees Forward Errors.} First, we evaluated all the approaches on knees inward (KIE) and forward (KFE) errors. The results are summarized in Table \ref{tab:res_ours_squat}. Additionally, here, we also considered a single-view, single-subject version of our cross-view, cross-subject pose-contrastive approach. In this version, anchor, positive, and negative inputs all belonged to the same video instance. We applied strong augmentations (rotation, translation, masking image regions, color channel order changing, zooming, blurring) during training this model. We refer to this approach as Vanilla-PC.
\begin{table}[]
\RawFloats
\begin{minipage}{.48\textwidth}
    \centering
    \resizebox{\textwidth}{!}{
\begin{tabular}{@{}lccc@{}}
\toprule
\multirow{2}{*}{\textbf{Feature extraction model}} & \multirow{2}{*}{\textbf{Modality}} & \multicolumn{2}{c}{\textbf{F-score $\uparrow$}} \\ \cmidrule(l){3-4} 
                                                   &                                    & \textbf{KIE}      & \textbf{KFE}      \\ \midrule
OpenPose-TDM \cite{openpose, ogata} & 2D Pose                             & 0.4143            & 0.8123                         \\
OpenPose-TDM$^*$ \cite{openpose, ogata} & 2D Pose                      & 0.3186            & 0.7968                         \\
SPIN-TDM \cite{spin, ogata}  & 3D Pose                              & 0.2878            & 0.7761                         \\ \midrule
ImageNet \cite{imagenet}  & Image                          & 0.1923            & 0.7725                         \\
SimSiam \cite{simsiam} & Image                            & 0.2270            & 0.7868                         \\
Ours PAD  & Image                           & 0.3180            & 0.7784                         \\
Ours Vanilla PC  & Image                    & 0.4118            & 0.7965                         \\
Ours CVCSPC  & Image                        & \textbf{0.5195}   & 0.8286                         \\ \midrule
Kinetics \cite{kinetics}  & Video                          & 0.2970            & 0.8184                         \\
VideoSpeed-1 \cite{benaim2020speednet}  & Video                          & 0.3095            & 0.8155                         \\
VideoSpeed-2   & Video                          & 0.3617            & 0.8000                         \\
VideoRot \cite{video_rotnet}  & Video                          &  0.3333           &  0.8138                        \\
TemporalXform \cite{jenni2020video}  & Video                          &  0.3414           & 0.8319                         \\
Ours TemporalXform-1   & Video                          & 0.3457            & 0.8097                         \\
Ours TemporalXform-2   & Video                          & 0.2286            & 0.8184                         \\
Ours MD  & Video                            & 0.4186            & \textbf{0.8338}                \\ \midrule
Ours MD + CVCSPC & Image, Video                    & \textbf{0.5263}   & \textbf{0.8468}                \\ \bottomrule
\end{tabular}
}
\caption{\textbf{Performance comparison on Knees Inward and Knees Forward errors on our BackSquat dataset}.}
\label{tab:res_ours_squat}
\end{minipage}%
\hfill
\begin{minipage}{.48\textwidth}
    \centering
    \resizebox{\textwidth}{!}{
\begin{tabular}{@{}lcc@{}}
\toprule
\textbf{Feature extraction model} & \textbf{Modality} & \textbf{F-score $\uparrow$} \\ \midrule
OpenPose-TDM \cite{openpose, ogata}  & 2D Pose       & 0.8340                 \\ \midrule
SimSiam \cite{simsiam} & Image & 0.8286 \\
Ours CVCSPC   & Image & \textbf{0.8694}                 \\ \bottomrule
\end{tabular}
}
\caption{\textbf{Performance comparison on detecting Shallow Squat error.}}
\label{tab:res_shallow_squat}
\end{minipage} 
\end{table}
We observed the following. 1) Training both image- and video-based self-supervision methods on our dataset helped in improving over their respective base models (ImageNet pretrained model and Kinetics pretrained model). 2) Our Vanilla Pose Contrastive learning improved the performance even more than our PAD. However, off-the-shelf pose estimator, OpenPose, still worked better than this model. 3) By contrast, our full pose-contrastive model, CVCSPC outperformed all the models on KIE; for completeness, we also computed OpenPose baseline with our hyperparameter settings referred to as OpenPose$^*$. 4) CVCSPC performing better than Vanilla PC also reinforced the importance of considering our cross-view and cross-subject conditions during pose contrastive learning. 5) Our MD model performed the best and second best on KFE and KIE, respectively. TemporalXform performed the best among general video self-supervised approaches. 6) Our domain knowledge-informed self-supervised approaches outperformed general self-supervised approaches, indicating the importance of using domain knowledge in designing self-supervised approaches. 7) Our contrastive learning-based approaches (CVCSPC and MD) worked better than our reconstruction-based approach (PAD). Furthermore, ensemble of our contrastive approaches outperformed all the models. Attention visualizations presented in the supplementary material.

Note that in all the subsequent experiments, we selected only the best performing methods for further evaluation.
\paragraph{Shallow Squat Error.} We further considered evaluating and comparing approaches on another squat error\textemdash shallow squat error. Since shallow depth error is a static type of error, image models (2DCNN-based) are more suitable, where errors are detected in singular images, as opposed to in a stack of video frames. Using a 3DCNN for detecting single frame-based errors does not make sense. Therefore, we have not considered our MD approach for single frame-based errors. Single image detection also made end-to-end learning more feasible, so we finetuned our models end-to-end. The results are summarized in Table \ref{tab:res_shallow_squat}. We observed that OpenPose worked better than SimSiam. Our self-supervised learning performed the best, showing the importance of learning task-oriented representations, and its utility even in end-to-end finetuning scenarios.
\subsubsection{Dataset: Fitness-AQA OverheadPress.}
Further, we evaluated and compared approaches on a different exercise\textemdash OverheadPress. The results are summarized in Table \ref{tab:res_ohp}. We observed that video-based approaches worked better than image-based approaches on this exercise. Both of our proposed approaches outperformed the off-the-shelf pose estimator.
\begin{table}[]
\RawFloats
\begin{minipage}{.48\textwidth}
    \centering
    \resizebox{\textwidth}{!}{
\begin{tabular}{@{}lccc@{}}
\toprule
\multirow{2}{*}{\textbf{Feature extraction model}} & \multirow{2}{*}{\textbf{Modality}} & \multicolumn{2}{c}{\textbf{F-score $\uparrow$}} \\ \cmidrule(l){3-4} 
                         &                          & \textbf{Elbow Err.}  & \textbf{Knees Err.}            \\ \midrule
OpenPose-TDM \cite{openpose, ogata} & 2D Pose    & 0.4265          & 0.7131          \\ \midrule
SimSiam \cite{simsiam} & Image  & 0.4145              & 0.5301 \\
Ours CVCSPC & Image & 0.4522          & 0.7203          \\ \midrule
TemporalXform \cite{jenni2020video}   & Video       & 0.4138          & 0.8416   \\ 
Ours MD & Video    & \textbf{0.4552} & \textbf{0.8452} \\ \bottomrule
\end{tabular}
}
\caption{\textbf{Performance comparison on detecting Elbow and Knees errors in OverheadPress exercise}.}
\label{tab:res_ohp}
\end{minipage}%
\hfill
\begin{minipage}{.48\textwidth}
    \centering
    \resizebox{\textwidth}{!}{
\begin{tabular}{@{}lccc@{}}
\toprule
\multirow{2}{*}{\textbf{Feature extraction model}} & \multirow{2}{*}{\textbf{Modality}} & \multicolumn{2}{c}{\textbf{F-score $\uparrow$}} \\ \cmidrule(l){3-4} 
                         &                           &   \textbf{Lumbar Err.}          & \textbf{Torso Err.}             \\ \midrule
OpenPose-TDM \cite{openpose, ogata} (SQ$\rightarrow$BR) & 2D Pose & 0.5422          & 0.4060          \\ \midrule
SimSiam \cite{simsiam} (SQ$\rightarrow$BR) & Image & 0.5934           & 0.4543 \\
Ours CVCSPC (SQ$\rightarrow$BR)  & Image & \textbf{0.6057} & \textbf{0.4800} \\
Ours CVCSPC (OHP$\rightarrow$BR)  & Image & 0.5760 & 0.4675 \\
Ours CVCSPC (SQ+OHP$\rightarrow$BR)  & Image & \textbf{0.6338} & \textbf{0.5261} \\
\bottomrule
\end{tabular}
}
\caption{\textbf{Cross-exercise transfer performance. Detecting Lumbar and Torso-Angle errors in BarbellRow exercise.}}
\label{tab:res_barbellrow}
\end{minipage} 
\end{table}
\subsection{Cross-Exercise Transfer}
\label{exp:cross_exercise_transfer}
It is common to not have enough labeled data for each exercise. In such cases, it would be useful to transfer models from an exercise with abundant data over to exercises with limited data. So, in this experiment, we first transferred our model trained on BackSquat (SQ) exercise to BarbellRow exercise, where we detected two kinds of errors: Lumbar and TorsoAngle errors. Since these errors are static errors, we considered transferring our CVCSPC model. Note that in this experiment we used only a small amount of training data (details in the Supplementary Material). The results are presented in Table \ref{tab:res_barbellrow}. We observed that models pretrained using our proposed self-supervised approach performed better than baselines even when finetuned to a different exercise action. We also transferred from Overhead Press (OHP), \& noted improvements. Lastly, we also tried the ensemble of our SQ \& OHP transferred models, which worked the best.
\subsection{Applications to Other Domains}
\subsubsection{Pose Estimation.} 
\begin{figure}
    \centering
    \includegraphics[width=\textwidth]{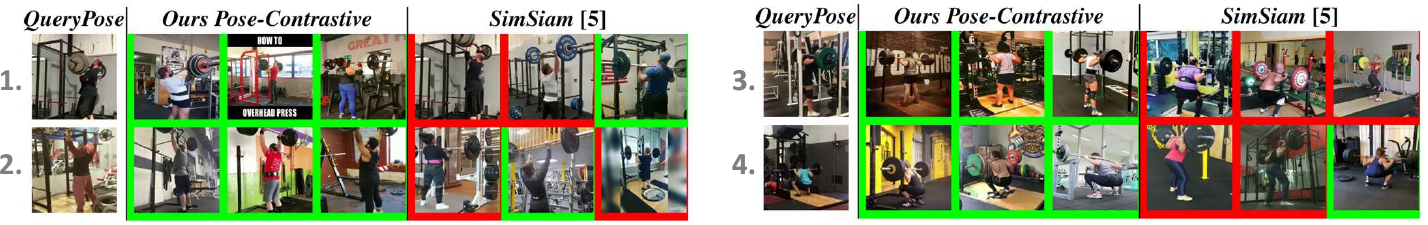}
    \caption{Results of \textbf{pose-based retrieval} experiment.}
    \label{fig:pose_ret_res}
\end{figure}

We conducted a novel pose retrieval experiment where we retrieved images based on query poses using our pose-contrastive embeddings. From the results shown in Fig. \ref{fig:pose_ret_res}, it can be seen that compared to SimSiam embeddings, ours are much better at encoding pose information, even with camera angle variation. We believe that our representations can be decoded into actual 2D/3D joint positions, by using a small pose-annotated dataset. We will explore this further in future research.

\subsubsection{Dive Quality Assessment.}
\textit{While we use symmetry to simplify problems, our methods are generalizable}, \eg, we applied our motion disentangling method for assessing the quality of Olympic dives on MTL-AQA dataset \cite{mtlaqa}. Global motion \& local motions here refer to the motion of the dive-classes \& the errors in them, respectively. To disentangle local motion, we match-contrast dives from the same dive-class from the same diving events so that the background remains same. We used supervised dive-classification pretraining as the baseline. Performance metric is Spearman's rank correlation (higher is better). We found significant improvement after incorporating our motion disentangling approach as shown in Table~\ref{tab:res_dive_aqa}, even surpassing previous self-supervised state-of-the-art \cite{roditakis2021towards}.
\begin{table}[]
\centering
\resizebox{0.3\textwidth}{!}{
\begin{tabular}{l|c|c|c}
\toprule
\textbf{Model} & \textbf{SSL SoTA} \cite{roditakis2021towards} & \textbf{Ours baseline} & \textbf{Ours MD} \\ \midrule
\textbf{Sp. Corr.}  & 0.7700 & 0.5665    & \textbf{0.7763}           \\ 
\bottomrule
\end{tabular}}
\caption{\textbf{Motion disentangling for Dive quality assessment.}}
\label{tab:res_dive_aqa}
\end{table}
\section{Conclusion}
In this paper, we addressed the problem of assessing the workout form in real-world gym scenarios, where we showed that pose-features from off-the-shelf pose estimators cannot be reliably used for detecting subtle errors in workout form, as these pose estimators struggle to perform well due to unusual poses, occlusions, illumination, and clothing styles. We tackled the problem by replacing these noisy pose features with our more robust image and video representations learnt from unlabeled videos using domain knowledge-informed self-supervised approaches. Using self-supervision helped in avoiding the cost of annotating poses. Mapping of our self-supervised representations to workout form error probabilities was learnt using a much smaller labeled dataset. We also introduced a novel dataset, Fitness-AQA, containing actual, unscripted exercise samples from real-world gyms. Experimentally, we found that while our self-supervised features performed comparably in simpler conditions, they outperformed off-the-shelf pose estimators and various baselines in complex real-world conditions on multiple exercises. We also showed that pose information is encoded in our representations; and our motion disentangling approach can be used to assess quality of motion in other domains.

\clearpage
%
%
\bibliographystyle{splncs04}
\bibliography{egbib}
\end{document}